\documentclass[twoside,11pt]{article}
\usepackage{jair, rawfonts}
\usepackage[utf8]{inputenc}
\usepackage[main=american]{babel}
\usepackage[babel=true]{csquotes}
\usepackage[parfill]{parskip}
\usepackage{booktabs}
\usepackage{multirow}
\usepackage{subcaption}
\usepackage{suffix}
\usepackage{xcolor}
\usepackage[natbibapa]{apacite}
\usepackage{tabularx} 
\usepackage{enumitem}
\usepackage{CJKutf8}
\usepackage[hidelinks,breaklinks]{hyperref}

\newcommand{\Section}[1]{Section~\ref{sec:#1}}
\newcommand{\Table}[1]{Table~\ref{tab:#1}}

\newcommand{\HA}{H$_A$ }
\WithSuffix\newcommand\HA*{H$_A$}
\newcommand{\HB}{H$_B$ }
\WithSuffix\newcommand\HB*{H$_B$}
\newcommand{\MT}{MT$_1$ }
\WithSuffix\newcommand\MT*{MT$_1$}
\newcommand{\GG}{MT$_2$ }
\WithSuffix\newcommand\GG*{MT$_2$}

\newcommand{\ie}{i.\,e.\ }
\WithSuffix\newcommand\ie*{i.\,e.}
\newcommand{\Ie}{I.\,e.\ }
\WithSuffix\newcommand\Ie*{I.\,e.}
\newcommand{\eg}{e.\,g.\ }
\WithSuffix\newcommand\eg*{e.\,g.}
\newcommand{\Eg}{E.\,g.\ }
\WithSuffix\newcommand\Eg*{E.\,g.}
\newcommand{\ibid}{ibid.\ }
\WithSuffix\newcommand\ibid*{ibid.}
\newcommand{\Ibid}{Ibid.\ }
\WithSuffix\newcommand\Ibid*{Ibid.}
\newcommand{\percent}{\,\%\ }
\WithSuffix\newcommand\percent*{\,\%}
\newcommand{\vs}{vs.\ }
\WithSuffix\newcommand\vs*{\vs.}

\newcommand{\citeg}[1]{\citeauthor{#1}'s \citeyearpar{#1}} 

\jairheading{67}{2020}{653--672}{02/2019}{03/2020}
\ShortHeadings{Assessing Human--Machine Parity in Language Translation}
{Läubli, Castilho, Neubig, Sennrich, Shen, \& Toral}
\firstpageno{653}

\title{A Set of Recommendations for Assessing \\Human--Machine Parity in Language Translation\bigskip}
\author{\name Samuel Läubli \email laeubli@cl.uzh.ch \\
        \addr Institute of Computational Linguistics, University of Zurich
        \AND
        \name Sheila Castilho \email sheila.castilho@adaptcentre.ie \\
        \addr ADAPT Centre, Dublin City University
        \AND 
        \name Graham Neubig \email gneubig@cs.cmu.edu \\
        \addr Language Technologies Institute, Carnegie Mellon University
        \AND 
        \name Rico Sennrich \email sennrich@cl.uzh.ch \\
        \addr Institute of Computational Linguistics, University of Zurich
        \AND 
        \name Qinlan Shen \email qinlans@cs.cmu.edu \\
        \addr Language Technologies Institute, Carnegie Mellon University
        \AND 
        \name Antonio Toral \email a.toral.ruiz@rug.nl \\
        \addr Center for Language and Cognition, University of Groningen}
        
\date{}

\begin{document}
\begin{CJK*}{UTF8}{gbsn}

\maketitle

\begin{abstract}
    The quality of machine translation has increased remarkably over the past years, to the degree that it was found to be indistinguishable from professional human translation in a number of empirical investigations. We reassess Hassan et al.'s 2018 investigation into Chinese to English news translation, showing that the finding of human--machine parity was owed to weaknesses in the evaluation design---which is currently considered best practice in the field. We show that the professional human translations contained significantly fewer errors, and that perceived quality in human evaluation depends on the choice of raters, the availability of linguistic context, and the creation of reference translations. Our results call for revisiting current best practices to assess strong machine translation systems in general and human--machine parity in particular, for which we offer a set of recommendations based on our empirical findings.
\end{abstract}

\section{Introduction}
\label{sec:Introduction}

Machine translation (MT) has made astounding progress in recent years thanks to improvements in neural modelling \citep{Sutskever2014,Bahdanau2014,NIPS2017_7181}, and the resulting increase in translation quality is creating new challenges for MT evaluation.
Human evaluation remains the gold standard, but there are many design decisions that potentially affect the validity of such a human evaluation.

This paper is a response to two recent human evaluation studies in which some neural machine translation systems reportedly performed at (or above) the level of human translators for news translation from Chinese to English \citep{hassan2018achieving} and English to Czech \citep{Popel2018,WMT2018}.

Both evaluations were based on current best practices in the field: they used a source-based direct assessment with non-expert annotators, using data sets and the evaluation protocol of the Conference on Machine Translation (WMT).
While the results are intriguing, especially because they are based on best practices in MT evaluation, \citet[p.~293]{WMT2018} warn against taking their results as evidence for human--machine parity, and caution that \enquote{for well-resourced language pairs, an update of WMT evaluation style will be needed to keep up with the progress in machine translation.}
We concur that these findings have demonstrated the need to critically re-evaluate the design of human MT evaluation.

Our paper investigates three aspects of human MT evaluation, with a special focus on assessing human--machine parity: the choice of raters, the use of linguistic context, and the creation of reference translations.
We focus on the data shared by \citet{hassan2018achieving}, and empirically test to what extent changes in the evaluation design affect the outcome of the human evaluation.\footnote{Our results synthesise and extend those reported by \citet{Laeubli2018} and \citet{Toral2018}.}
We find that for all three aspects, human translations are judged more favourably, and significantly better than MT, when we make changes that we believe strengthen the evaluation design.
Based on our empirical findings, we formulate a set of recommendations for human MT evaluation in general, and assessing human--machine parity in particular.
All of our data are made publicly available for external validation and further analysis.\footnote{\url{https://github.com/ZurichNLP/mt-parity-assessment-data}}

\section{Background}
\label{sec:Background}

We first review current methods to assess the quality of machine translation system outputs, and highlight potential issues in using these methods to compare such outputs to translations produced by professional human translators.

\subsection{Human Evaluation of Machine Translation}
\label{sec:BackgroundEvaluation}

The evaluation of MT quality has been the subject of controversial discussions in research and the language services industry for decades due to its high economic importance. While automatic evaluation methods are particularly important in system development, there is consensus that a reliable evaluation should---despite high costs---be carried out by humans.

Various methods have been proposed for the human evaluation of MT quality \citep[c.f.][]{CASTILHO_2018_TQA}. What they have in common is that the MT output to be rated is paired with a translation hint: the source text or a reference translation. The MT output is then either adapted or scored with reference to the translation hint by human post-editors or raters, respectively.

As part of the large-scale evaluation campaign at WMT, two primary evaluation methods have been used in recent years: relative ranking and direct assessment \citep{bojar-etal_Cracker:2016}. In the case of relative ranking, raters are presented with outputs from two or more systems, which they are asked to evaluate relative to each other (e.g., to determine system A is better than system B). Ties (e.g., system A is as good or as bad as system B) are typically allowed. Compared to absolute scores on Likert scales, data obtained through relative ranking show better inter- and intra-annotator agreement \citep{CallisonBurch2007}. However, they do not allow conclusions to be drawn about the order of magnitude of the differences, so that it is not possible to determine \textit{how much} better system A was than system B.

This is one of the reasons why direct assessment has prevailed as an evaluation method more recently. In contrast to relative ranking, the raters are presented with one MT output at a time, to which they assign a score between 0 and 100. To increase homogeneity, each rater's ratings are standardised \citep{Graham2013}. Reference translations serve as the basis in the context of WMT, and evaluations are carried out by monolingual raters.
To avoid reference bias, the evaluation can be based on source texts instead, which presupposes bilingual raters, but leads to more reliable results overall \citep{Bentivogli2018}.

\subsection{Assessing Human--Machine Parity}
\label{sec:BackgroundParity}

\citet{hassan2018achieving} base their claim of achieving human--machine parity on a source-based direct assessment as described in the previous section, where they found no significant difference in ratings between the output of their MT system and a professional human translation.
Similarly, \citet{WMT2018} report that the best-performing English to Czech system submitted to WMT 2018 \citep{Popel2018} significantly outperforms the human reference translation.
However, the authors caution against interpreting their results as evidence of human--machine parity, highlighting potential limitations of the evaluation.

In this study, we address three aspects that we consider to be particularly relevant for human evaluation of MT, with a special focus on testing human--machine parity: the choice of raters, the use of linguistic context, and the construction of reference translations.

\paragraph{Choice of Raters}
The human evaluation of MT output in research scenarios is typically conducted by crowd workers in order to minimise costs.
\citet{CallisonBurch2009} shows that aggregated assessments of bilingual crowd workers are \enquote{very similar} to those of MT developers, and \citet{Graham2017}, based on experiments with data from WMT 2012, similarly conclude that with proper quality control, MT systems can be evaluated by crowd workers. 
\citet{hassan2018achieving} also use bilingual crowd workers, but the studies supporting the use of crowdsourcing for MT evaluation were performed with older MT systems, and their findings may not carry over to the evaluation of contemporary higher-quality neural machine translation (NMT) systems.
In addition, the MT developers to which crowd workers were compared are usually not professional translators.
We hypothesise that expert translators will provide more nuanced ratings than non-experts, and that their ratings will show a higher difference between MT outputs and human translations.

\paragraph{Linguistic Context}
MT has been evaluated almost exclusively at the sentence level, owing to the fact that most MT systems do not yet take context across sentence boundaries into account. However, when machine translations are compared to those of professional translators, the omission of linguistic context---\eg*, by random ordering of the sentences to be evaluated---does not do justice to humans who, in contrast to most MT systems, can and do take inter-sentential context into account \citep{VoigtJurafsky2012,Wang2017}. We hypothesise that an evaluation of sentences in isolation, as applied by \citet{hassan2018achieving}, precludes raters from detecting translation errors that become apparent only when inter-sentential context is available, and that they will judge MT quality less favourably when evaluating full documents.

\paragraph{Reference Translations}
The human reference translations with which machine translations are compared within the scope of a human--machine parity assessment play an important role. \citet{hassan2018achieving} used all source texts of the WMT 2017 Chinese--English test set for their experiments, of which only half were originally written in Chinese; the other half were translated from English into Chinese. Since translated texts are usually simpler than their original counterparts \citep{Laviosa1998b}, they should be easier to translate for MT systems. Moreover, different human translations of the same source text sometimes show considerable differences in quality, and a comparison with an MT system only makes sense if the human reference translations are of high quality. \citet{hassan2018achieving}, for example, had the WMT source texts re-translated as they were not convinced of the quality of the human translations in the test set. At WMT 2018, the organisers themselves noted that \enquote{the manual evaluation included several reports of ill-formed reference translations} \citep[p.~292]{WMT2018}.
We hypothesise that the quality of the human translations has a significant effect on findings of human--machine parity, which would indicate that it is necessary to ensure that human translations used to assess parity claims need to be carefully vetted for their quality.

We empirically test and discuss the impact of these factors on human evaluation of MT in Sections \ref{sec:Raters}--\ref{sec:ReferenceTranslations}.
Based on our findings, we then distil a set of recommendations for human evaluation of strong MT systems, with a focus on assessing human--machine parity (\Section{Recommendations}).

\subsection{Translations}
\label{sec:Translations}

We use English translations of the Chinese source texts in the WMT 2017 English--Chinese test set \citep{Bojar2017} for all experiments presented in this article:

\begin{description}[labelwidth=1cm, leftmargin=1.25cm]
    \item[\HA~~] The professional human translations in the dataset of \citet{hassan2018achieving}.\footnotemark[1]
    \item[\HB~~] Professional human translations that we ordered from a different translation vendor, which included a post-hoc native English check. We produced these only for the documents that were originally Chinese, as discussed in more detail in Section \ref{sec:Directionality}.
    \item[\MT] The machine translations produced by \citeg{hassan2018achieving} best system (\textsc{Combo-6}),\footnotemark[1] for which the authors found parity with \HA*.
    \item[\GG] The machine translations produced by Google's production system (Google Translate) in October 2017, as contained in \citeg{hassan2018achieving} dataset.\footnotemark[1]
\end{description}

Statistical significance is denoted by * ($p\le.05$), ** ($p\le.01$), and *** ($p\le.001$) throughout this article, unless otherwise stated.

\section{Choice of Raters}
\label{sec:Raters}

Both professional and amateur evaluators can be involved in human evaluation of MT quality. However, from  published work in the field \citep{Doherty_2017}, it is fair to say that there is a tendency to ``rely on students and amateur evaluators, sometimes with an undefined (or self-rated) proficiency in the languages involved, an unknown expertise with the text type" \citep[p.~23]{CASTILHO_2018_TQA}. 

Previous work on evaluation of MT output by professional translators against crowd workers by \citet{Castilho2017crowdsourcing} showed that for all language pairs (involving 11 languages) evaluated, crowd workers tend to be more accepting of the MT output by giving higher fluency and adequacy scores and performing very little post-editing. The authors argued that non-expert translators lack knowledge of translation and so might not notice subtle differences that make one translation more suitable than another, and therefore, when confronted with a translation that is hard to post-edit, tend to accept the MT rather than try to improve it. 

\subsection{Evaluation Protocol}

We test for difference in ratings of MT outputs and human translations between experts 
and non-experts. We consider professional translators as experts, and both crowd workers and MT researchers as non-experts.\footnote{This terminology is not consistent with other literature, where MT researchers have been referred to as experts and crowd workers as non-experts \citep[\eg*,][]{CallisonBurch2009}.} 

We conduct a relative ranking experiment using one professional human (\HA*) and two machine translations (\MT and \GG*), considering the native Chinese part of the WMT 2017 Chinese--English test set (see \Section{Directionality} for details).
The 299 sentences used in the experiments stem from 41 documents, randomly selected from all the documents in the test set originally written in Chinese, and are shown in their original order.
Raters are shown one sentence at a time, and see the original Chinese source alongside the three translations.
The previous and next source sentences are also shown, in order to provide the annotator with local inter-sentential context.

Five raters---two experts and three non-experts---participated in the assessment. The experts were professional Chinese to English translators: one native in Chinese with a fluent level of English, the other native in English with a fluent level of Chinese. The non-experts were NLP researchers native in Chinese, working in an English-speaking country.

The ratings are elicited with Appraise \citep{mtm12_appraise}. We derive an overall score for each translation (\HA*, \MT*, and \GG*) based on the rankings.  We use the TrueSkill method adapted to MT evaluation~\citep{sakaguchi-post-vandurme:2014:W14-33} 
following its usage at WMT15,\footnote{\url{https://github.com/mjpost/wmt15}}
\ie*, we run 1,000 iterations of the rankings recorded with Appraise followed by clustering (significance level $\alpha=0.05$).

\subsection{Results}

\begin{table}
    \centering
    \begin{tabular}{llrlrlr}
\toprule
\textbf{Rank} & \multicolumn{6}{l}{\textbf{Translators}}                                                                                                                                                                                                   \\
              & \multicolumn{2}{l}{All} & \multicolumn{2}{l}{Experts} & \multicolumn{2}{l}{Non-experts} \\[-0.5em]
              & \multicolumn{2}{l}{$n=3873$} & \multicolumn{2}{l}{$n=1785$} & \multicolumn{2}{l}{$n=2088$} \\ 
              \hline
1             & \HA                               & $1.939$~*                              & \HA                                 & $2.247$~*                                & \HA                                   & $1.324$~\phantom{*}                                   \\
2             & \MT                               & $1.199$~*                              & \MT                                 & $1.197$~*                                & \MT                                   & $0.940$~*                                   \\
3             & \GG                               & $-3.144$~\phantom{*}                              & \GG                                 & $-3.461$~\phantom{*}                                & \GG                                   & $-2.268$~\phantom{*}                                  \\ \bottomrule
\end{tabular}
    \smallskip
    \caption{Ranks and TrueSkill scores (the higher the better) of one human (\HA*) and two machine translations (\MT*, \GG*) for evaluations carried out by expert and non-expert translators. An asterisk next to a translation indicates that this translation is significantly better than the one in the next rank at $p\leq.05$.}
    \label{t:evaluators}
\end{table}

Table \ref{t:evaluators} shows the TrueSkill scores for each translation resulting from the evaluations by expert and non-expert translators. We find that translation expertise affects the judgement of \MT and \HA*, where the rating gap is wider 
for the expert raters.\footnote{As mentioned before, relative ranking mostly tells whether a translation is better than another but not by how much. The TrueSkill score is able to measure that difference, but may be difficult to interpret.} This indicates that non-experts disregard translation nuances in the evaluation, which leads to a more tolerant judgement of MT systems and a lower inter-annotator agreement ($\kappa=0.13$ for non-experts versus $\kappa=0.254$ for experts). 

It is worth noticing that, regardless of their expertise, the performance of human raters may vary over time. For example, performance may improve or decrease due to learning effects or fatigue, respectively \citep{GONZALEZ201119}. It is likely that such longitudinal effects are present in our data. They should be accounted for in future work, \eg*, by using trial number as an additional predictor~\citep{toral_penovel_18}.

\section{Linguistic Context}
\label{sec:Context}

Another concern is the unit of evaluation. Historically, machine translation has primarily operated on the level of sentences, and so has machine translation evaluation. However, it has been remarked that human raters do not necessarily understand the intended meaning of a sentence shown out-of-context \citep{Wu2016}, which limits their ability to spot some mistranslations. Also, a sentence-level evaluation will be blind to errors related to textual cohesion and coherence.

While sentence-level evaluation may be good enough when evaluating MT systems of relatively low quality, we hypothesise that with additional context, raters will be able to make more nuanced quality assessments, and will also reward translations that show more textual cohesion and coherence. We believe that this aspect should be considered in evaluation, especially when making claims about human--machine parity, since human translators can and do take inter-sentential context into account \citep{VoigtJurafsky2012,Wang2017}.

\subsection{Evaluation Protocol}
\label{sec:ContextEvaluationProtocol}

We test if the availability of document-level context affects human--machine parity claims in terms of adequacy and fluency. In a pairwise ranking experiment, we show raters (i) isolated sentences and (ii) entire documents, asking them to choose the better (with ties allowed) from two translation outputs: one produced by a professional translator, the other by a machine translation system. We do not show reference translations as one of the two options is itself a human translation.

We use source sentences and documents from the WMT 2017 Chinese--English test set (see \Section{Translations}): documents are full news articles, and sentences are randomly drawn from these news articles, regardless of their position. We only consider articles from the test set that are native Chinese (see \Section{Directionality}). In order to compare our results to those of \citet{hassan2018achieving}, we use both their professional human (\HA*) and machine translations (\MT*). 

Each rater evaluates both sentences and documents, but never the same text in both conditions so as to avoid repetition priming \citep{FrancisSaenz2007}. The order of experimental items as well as the placement of choices (\HA*, \MT*; left, right) are randomised.

We use spam items for quality control \citep{Kittur2008}: In a small fraction of items, we render one of the two options nonsensical by randomly shuffling the order of all translated words, except for 10\percent at the beginning and end. If a rater marks a spam item as better than or equal to an actual translation, this is a strong indication that they did not read both options carefully.

We recruit professional translators (see \Section{Raters}) from \url{proz.com}, a well-known online market place for professional freelance translation, considering Chinese to English translators and native English revisers for the adequacy and fluency conditions, respectively. In each condition, four raters evaluate 50 documents (plus 5 spam items) and 104 sentences (plus 16 spam items). We use two non-overlapping sets of documents and two non-overlapping sets of sentences, and each is evaluated by two raters.

\subsection{Results}
\label{sec:ContextResults}
\vspace{-1mm}

\begin{table}
    \begin{tabularx}{\textwidth}{@{}llp{2mm}crrXp{2mm}crrX@{}}
\toprule
\textbf{Context} & \textbf{N} &  & \multicolumn{4}{l}{\textbf{Adequacy}}                                                   &  & \multicolumn{4}{l}{\textbf{Fluency}}                                                      \\
                 &            &  & \multicolumn{1}{l}{\MT}          & \multicolumn{1}{l}{Tie} & \multicolumn{1}{l}{\HA} & $p$ &  & \multicolumn{1}{l}{\MT}                          & \multicolumn{1}{l}{Tie} & \multicolumn{1}{l}{\HA} & $p$   \\ \cmidrule(r){1-2} \cmidrule(lr){4-7} \cmidrule(l){9-12} 
Sentence         & 208        &  & \multicolumn{1}{r}{49.5\percent} & 9.1\percent                  & 41.4\percent                     &   &  & \multicolumn{1}{r}{31.7\percent} & 17.3\percent                 & 51.0\percent                     & **  \\
Document         & 200        &  & \multicolumn{1}{r}{37.0\percent} & 11.0\percent                 & 52.0\percent                     & * &  & \multicolumn{1}{r}{22.0\percent} & 28.5\percent                 & 49.5\percent                     & *** \\ \bottomrule
\end{tabularx}
    \smallskip
    \caption{Pairwise ranking results for machine (\MT*) against professional human translation (\HA*) as obtained from blind evaluation by professional translators. Preference for \MT is lower when document-level context is available.}
    \label{tab:mt-vs-a}
\end{table}

Results are shown in \Table{mt-vs-a}. We note that sentence ratings from two raters are excluded from our analysis because of unintentional textual overlap with documents, meaning we cannot fully rule out that sentence-level decisions were informed by access to the full documents they originated from. Moreover, we exclude document ratings from one rater in the fluency condition because of poor performance on spam items, and recruit an additional rater to re-rate these documents.

We analyse our data using two-tailed Sign Tests, the null hypothesis being that raters do not prefer \MT over \HA or vice versa, implying human--machine parity. Following WMT evaluation campaigns that used pairwise ranking \citep[\eg*,][]{WMT2013}, the number of successes $x$ is the number of ratings in favour of \HA*, and the number of trials $n$ is the number of all ratings except for ties. Adding half of the ties to $x$ and the total number of ties to $n$ \citep{EmersonSimon1979} does not impact the significance levels reported in this section.

Adequacy raters show no statistically significant preference for \MT or \HA when evaluating isolated sentences ($x=86, n=189, p=.244$). This is in accordance with \citet{hassan2018achieving}, who found the same in a source-based direct assessment experiment with crowd workers. With the availability of document-level context, however, preference for \MT drops from 49.5 to 37.0\percent and is significantly lower than preference for human translation ($x=104, n=178, p<.05$). This evidences that document-level context cues allow raters to get a signal on adequacy.

Fluency raters prefer \HA over \MT both on the level of sentences ($x=106, n=172, p<.01$) and documents ($x=99, n=143, p<.001$). This is somewhat surprising given that increased fluency was found to be one of the main strengths of NMT \citep{WMT2016}, as we further discuss in \Section{Quality}. The availability of document-level context decreases fluency raters' preference for \MT*, which falls from 31.7 to 22.0\percent*, without increasing their preference for \HA  (\Table{mt-vs-a}).

\subsection{Discussion}

\begin{table}
    \begin{tabularx}{\textwidth}{p{1cm}X}
    \hline
    Source & 传统习俗引入新亮点``\textbf{2016盂兰文化节}"香港维园开幕 敲锣打鼓的音乐、传统的小食、花俏的装饰、人群汹涌的现场。 由香港潮属社团总会主办的``\textbf{2016盂兰文化节}"12日至14日在维多利亚公园举办，这是香港最盛大的一场盂兰胜会。\\
    \HA     & Traditional customs with new highlights - \textbf{2016 Ullam Cultural Festival} unveiled at Victoria Park in Hong Kong Music with drums and gongs, traditional snacks, fanciful decorations, and a chock-a-block crowd at the scene. The ``\textbf{2016 Ullam Cultural Festival}'' organized by the Federation of Hong Kong Chiu Chow Community Organizations will be held at Victoria Park from 12th to the 14th. \\
    \MT     & Traditional customs introduce new bright spot ``\textbf{2016 Ullambana Cultural Festival}'' Hong Kong Victoria Park opening Gongs and drums music, traditional snacks, fancy decorations, the crowd surging scene. Organised by the Federation of Teochew Societies in Hong Kong, the ``\textbf{2016 Python Cultural Festival}'' is held at Victoria Park from 12 to 14 July. \\
    \hline
\end{tabularx}
    \smallskip
    \caption{Two consecutive sentences of a Chinese news article as translated into English by a professional human translator (\HA*) and a machine translation system (\MT*). Emphasis added.}
    \label{tab:context-example}
\end{table}

Our findings emphasise the importance of linguistic context in human evaluation of MT. In terms of adequacy, raters assessing documents as a whole show a significant preference for human translation, but when assessing single sentences in random order, they show no significant preference for human translation.

Document-level evaluation exposes errors to raters which are hard or impossible to spot in a sentence-level evaluation, such as coherent translation of named entities. The example in \Table{context-example} shows the first two sentences of a Chinese news article as translated by a professional human translator (\HA*) and \citeg{hassan2018achieving} NMT system (\MT*). When looking at both sentences (document-level evaluation), it can be seen that \MT uses two different translations to refer to a cultural festival, ``2016盂兰文化节", whereas the human translation uses only one. When assessing the second sentence out of context (sentence-level evaluation), it is hard to penalise \MT for producing \enquote{2016 Python Cultural Festival}, particularly for fluency raters without access to the corresponding source text. For further examples, see \Section{Quality} and \Table{quality_qualitative}.

\section{Reference Translations}
\label{sec:ReferenceTranslations}

Yet another relevant element in human evaluation is the reference translation used.
This is the focus of this section, where we cover two aspects of reference translations that can have an impact on evaluation: quality and directionality.

\subsection{Quality}
\label{sec:Quality}

Because the translations are created by humans, a number of factors could lead to compromises in quality:
\begin{description}
\item[Errors in Understanding:] If the translator is a non-native speaker of the source language, they may make mistakes in interpreting the original message. This is particularly true if the translator does not normally work in the domain of the text, \eg*, when a translator who normally works on translating electronic product manuals is asked to translate news.
\item[Errors in Fluency:] If the translator is a non-native speaker of the target language, they might not be able to generate completely fluent text. This similarly applies to domain-specific terminology.
\item[Limited Resources:] Unlike computers, human translators have limits in time, attention, and motivation, and will generally do a better job when they have sufficient time to check their work, or are particularly motivated to do a good job, such as when doing a good job is necessary to maintain their reputation as a translator.
\item[Effects of Post-editing:] In recent years, a large number of human translation jobs are performed by post-editing MT output, which can result in MT artefacts remaining even after manual post-editing \citep{daems2017translation,toral-2019-post,castilho-etal-2019-influences}.
\end{description}

In this section, we examine the effect of the quality of underlying translations on the conclusions that can be drawn with regards to human--machine parity.
We first do an analysis on (i) how the source of the human translation affects claims of human--machine parity, and (ii) whether significant differences exist between two varieties of human translation.
We follow the same protocol as in \Section{ContextEvaluationProtocol}, having 4 professional translators per condition, evaluate the translations for adequacy and fluency on both the sentence and document level.\footnote{Translators were recruited from \url{proz.com}.}

\begin{table}
    \begin{subtable}[b]{\textwidth}
        \begin{tabularx}{\textwidth}{@{}llp{2mm}crrXp{2mm}crrX@{}}
\toprule
\textbf{Context} & \textbf{N} &  & \multicolumn{4}{l}{\textbf{Adequacy}}                                                   &  & \multicolumn{4}{l}{\textbf{Fluency}}                                                      \\
                 &            &  & \multicolumn{1}{l}{\MT}                          & \multicolumn{1}{l}{Tie} & \multicolumn{1}{l}{\HB} & $p$ &  & \multicolumn{1}{l}{\MT}                          & \multicolumn{1}{l}{Tie} & \multicolumn{1}{l}{\HB} & $p$   \\ \cmidrule(r){1-2} \cmidrule(lr){4-7} \cmidrule(l){9-12} 
Sentence         & 416        &  & \multicolumn{1}{r}{34.6\percent} & 18.8\percent                 & 46.6\percent                     &  &  & \multicolumn{1}{r}{20.7\percent} & 21.2\percent                 & 58.2\percent                     & *** \\
Document         & 200        &  & \multicolumn{1}{r}{41.0\percent} & 9.0\percent                  & 50.0\percent                     & \phantom{***}   &  & \multicolumn{1}{r}{18.0\percent} & 21.0\percent                 & 61.0\percent                     & *** \\[-1em] \bottomrule
\end{tabularx}
        \smallskip
        \caption{Machine translation \MT against professional human translation \HB}
        \label{tab:mt-vs-b}
        \medskip
    \end{subtable}
    \begin{subtable}[b]{\textwidth}
        \begin{tabularx}{\textwidth}{@{}llp{2mm}crrXp{2mm}crrX@{}}
\toprule
\textbf{Context} & \textbf{N} &  & \multicolumn{4}{l}{\textbf{Adequacy}}                                                     &  & \multicolumn{4}{l}{\textbf{Fluency}}                                                    \\
                 &            &  & \multicolumn{1}{l}{\HA}                     & \multicolumn{1}{l}{Tie} & \multicolumn{1}{l}{\HB} & $p$   &  & \multicolumn{1}{l}{\HA}                    & \multicolumn{1}{l}{Tie} & \multicolumn{1}{l}{\HB} & $p$ \\ \cmidrule(r){1-2} \cmidrule(lr){4-7} \cmidrule(l){9-12} 
Sentence         & 416        &  & \multicolumn{1}{r}{56.7 \%} & 10.6 \%                 & 32.7 \%                     & *** &  & \multicolumn{1}{r}{40.4 \%} & 24.0 \%                 & 35.6 \%                     &   \\
Document         & 200        &  & \multicolumn{1}{r}{64.0 \%} & 9.0 \%                  & 27.0 \%                     & *** &  & \multicolumn{1}{r}{34.0 \%} & 22.0 \%                 & 44.0 \%                     & \phantom{***}   \\[-1em] \bottomrule
\end{tabularx}
        \smallskip
        \caption{Professional human translation \HA against professional human translation \HB}
        \label{tab:a-vs-b}
    \end{subtable}
    \caption{Pairwise ranking results for one machine (\MT*) and two professional human translations (\HA*, \HB*) as obtained from blind evaluation by professional translators.}
    \label{tab:quality_main}
\end{table}

The results are shown in Table \ref{tab:quality_main}.
From this, we can see that the human translation \HB*, which was aggressively edited to ensure target fluency, resulted in lower adequacy (\Table{a-vs-b}).
With more fluent and less accurate translations, raters do not prefer human over machine translation in terms of adequacy (\Table{mt-vs-b}), but have a stronger preference for human translation in terms of fluency (compare Tables~\ref{tab:mt-vs-b} and~\ref{tab:mt-vs-a}).
In a direct comparison of the two human translations (\Table{a-vs-b}), we also find that \HA is considered significantly more adequate than \HB*, while there is no significant difference in fluency.

To achieve a finer-grained understanding of what errors the evaluated translations exhibit, we perform a categorisation of 150 randomly sampled sentences based on the classification used by \citet{hassan2018achieving}.\footnote{\citeg{hassan2018achieving} classification is in turn based on, but significantly different than that proposed by \citet{vilar2006error}.}
We expand the classification with a Context category, which we use to mark errors that are only apparent in larger context (\eg*, regarding poor register choice, or coreference errors), and which do not clearly fit into one of the other categories.
\citet{hassan2018achieving} perform this classification only for the machine-translated outputs, and thus the natural question of whether the mistakes that humans and computers make are qualitatively different is left unanswered.
Our analysis was performed by one of the co-authors who is a bi-lingual native Chinese/English speaker.
Sentences were shown in the context of the document, to make it easier to determine whether the translations were correct based on the context.
The analysis was performed on one machine translation (\MT*) and two human translation outputs (\HA*, \HB*), using the same 150 sentences, but blinding their origin by randomising the order in which the documents were presented.
We show the results of this analysis in Table \ref{tab:quality_error_classification}.

\begin{table}
    \centering
    \small
\begin{tabularx}{\textwidth}{Xllllll}
\toprule
\textbf{Error Category}               & \multicolumn{3}{l}{\textbf{Errors}} & \multicolumn{3}{l}{\textbf{Significance}} \\
                                      & \HA        & \HB       & \MT       & \HA--~\HB    & \HA--~\MT    & \HB--~\MT   \\
\hline
Incorrect Word                        & 51         & 52        & 85        &              & ***          & ***         \\
\quad Semantics                       & 33         & 36        & 48        &              &             &             \\
\quad Grammaticality                  & 18         & 16        & 37        &              & **           & **          \\
Missing Word                          & 37         & 69        & 56        & ***          & *            &             \\
\quad Semantics                       & 22         & 62        & 34        & ***          &              & ***         \\
\quad Grammaticality                  & 15         & 7         & 22        &              &              & **          \\
Named Entity                          & 16         & 19        & 30        &              & *            &             \\
\quad Person                          & 1          & 10        & 10        & *            & *            &             \\
\quad Location                        & 5          & 4         & 6         &              &              &             \\
\quad Organization                    & 4          & 4         & 8         &              &              &             \\
\quad Event                           & 1          & 1         & 3         &              &              &             \\
\quad Other                           & 5          & 1         & 7         &              &              &            \\
Word Order                            & 1          & 4         & 17        &              & ***          & **          \\
Factoid                               & 1          & 1         & 6         &              &              &             \\
Word Repetition                       & 2          & 4         & 4         &              &              &             \\
Collocation                           & 15         & 18        & 27        &              &             &             \\
Unknown Words/Misspellings            & 0          & 1         & 0         &              &              &             \\
Context (Register, Coreference, etc.) & 6          & 9         & 12        &              &              &             \\
\hline
Any                                   & 81         & 103       & 118       & *            & ***          &            \\
Total                                 & 129        & 177       & 237       & **           & ***          & **          \\
\bottomrule
\end{tabularx}
    \caption{Classification of errors in machine translation \MT* and two professional human translation outputs \HA and \HB*. Errors represent the number of sentences (out of $N=150$) that contain at least one error of the respective type. We also report the number of sentences that contain at least one error of any category (Any), and the total number of error categories present in all sentences (Total). Statistical significance is assessed with Fisher's exact test (two-tailed) for each pair of translation outputs.}
    \label{tab:quality_error_classification}
\end{table}

\begin{table}
    \centering
    \begin{subtable}[b]{\textwidth}
        \begin{tabularx}{\textwidth}{p{1cm}X}
    \hline
    Source  & 在目前较为主流的\textbf{观点}中，番薯的引进主要有三条\textbf{线路}。\\
    \HA     & Currently more mainstream \textbf{perspectives} point to three \textbf{channels} for the introduction of sweet potatoes. \\
    \HB     & There are currently three \textbf{theories} in regards to how the sweet potato was introduced. \\
    \MT     & In the current more mainstream \textbf{point of view}, the introduction of sweet potato has three main \textbf{lines}.  \\
    \hline
\end{tabularx}

        \caption{Incorrect Word}
        \label{tab:incorrect_word}
        \bigskip
    \end{subtable}
    \begin{subtable}[b]{\textwidth}
        \begin{tabularx}{\textwidth}{p{1cm}X}
    \hline
    Source & 该企业位于青岛老城区的厂区\textbf{去年年底}全面\textbf{停产},环保搬迁至平度的新厂区。\\
    \HA     & This corporation, situated in the factory area of Old Town of Qingdao, \textbf{stopped its production} lines \textbf{at the end of last year}. \\
    \HB     & The same company had a plant in Qingdao's old town but was \textbf{shut down} \textbf{last year}. \\
    \MT     & The enterprise is located in the old city of Qingdao plant \textbf{at the end of last year} to \textbf{stop production}. \\
    \hline
\end{tabularx}

        \caption{Reordering}
        \label{tab:reordering}
        \bigskip
    \end{subtable}
    \begin{subtable}[b]{\textwidth}
        \begin{tabularx}{\textwidth}{p{1cm}X}
    \hline
    Source & 据知情人士透露，\textbf{近期}，苏宁高层与苹果公司频频见面，目的就是为了准备充足的货源。\\
    \HA     & Insider disclosure revealed that the upper management of Suning and Apple has met frequently \textbf{in recent times} for the purpose of preparing enough resources. \\
    \HB     & According to informed sources, Suning executives met frequently with Apple in order to prepare enough stock. \\
    \MT     & According to people familiar with the matter, \textbf{recently}, Suning executives have been meeting with Apple frequently in order to prepare an adequate supply. \\
    \hline
\end{tabularx}

        \caption{Missing Word (Semantics)}
        \label{tab:missing_words}
        \bigskip
    \end{subtable}
    \caption{Qualitative examples of differences in types of errors found in machine (\MT*) and two professional human translations (\HA*, \HB*). Emphasis added.}
\end{table}
\begin{table}
\ContinuedFloat
    \begin{subtable}[b]{\textwidth}
        \begin{tabularx}{\textwidth}{p{1cm}X}
    \hline
    Source & 张彬彬和\textbf{家人}聚少离多 ... \textbf{父母}说 ... \textbf{张彬彬}很少说\textbf{自己的辛苦}，更多的是跟\textbf{父母}聊些开心的事。\\
    \HA     & Zhang Binbin spends little time with \textbf{family}... \textbf{Her parents} said... Zhang Binbin seldom said \textbf{she found things difficult}. More often, \textbf{she} would chat about happy things with parents. \\
    \HB     & Zhang Binbin saw \textbf{her family} less... \textbf{Her parents} said... \textbf{she} would seldom talk about \textbf{her hardship} and would mostly talk about something happy with \textbf{her} parents \\
    \MT     & Zhang Binbin and \textbf{his family} gathered less... Parents said... Zhang Binbin rarely said \textbf{their hard work}, more with \textbf{their parents} to talk about something happy. \\
    \hline
\end{tabularx}

        \caption{Context}
        \label{tab:context}
    \end{subtable}
    \caption{(Continued from previous page.)}
    \label{tab:quality_qualitative}
\end{table}

From these results, we can glean a few interesting insights.
First, we find significantly larger numbers of errors of the categories of 
Incorrect Word and Named Entity in \MT*, indicating that the MT system is less effective at choosing correct translations for individual words than the human translators.
An example of this can be found in Table \ref{tab:incorrect_word}, where we see that the MT system refers to a singular ``point of view" and translates \enquote{线路} (channel, route, path) into the semantically similar but inadequate \enquote{lines}.
Interestingly, \MT has significantly more Word Order errors, one example of this being shown in Table \ref{tab:reordering}, with the relative placements of \enquote{at the end of last year} (去年年底) and \enquote{stop production} (停产).
This result is particularly notable given previous reports that NMT systems have led to great increases in reordering accuracy compared to previous statistical MT systems \citep{neubig15wat,bentivogli16neuralvsphrasebased}, demonstrating that the problem of generating correctly ordered output is far from solved even in very strong NMT systems.
Moreover, \HB had significantly more Missing Word (Semantics) errors than both \HA ($p<.001$) and \MT ($p<.001$), an indication that the proofreading process resulted in drops of content in favour of fluency.
An example of this is shown in Table \ref{tab:missing_words}, where \HB dropped the information that the meetings between Suning and Apple were \textit{recently} (近期) held.
Finally, while there was not a significant difference, likely due to the small number of examples overall, it is noticeable that \MT had a higher percentage of Collocation and Context errors, which indicate that the system has more trouble translating words that are dependent on longer-range context. 
Similarly, some Named Entity errors are also attributable to translation inconsistencies due to lack of longer-range context. 
Table \ref{tab:context} shows an example where we see that the MT system was unable to maintain a consistently gendered or correct pronoun for the female Olympic shooter Zhang Binbin (张彬彬).

Apart from showing qualitative differences between the three translations, the analysis also supports the finding of the pairwise ranking study: \HA is both preferred over \MT in the pairwise ranking study, and exhibits fewer translation errors in our error classification. \HB has a substantially higher number of missing words than the other two translations, which agrees with the lower perceived adequacy in the pairwise ranking.

However, the analysis not only supports the findings of the pairwise ranking study, but also adds nuance to it. Even though \HB has the highest number of deletions, and does worse than the other two translations in a pairwise adequacy ranking, it is similar to \HA*, and better than \MT*, in terms of most other error categories.

\subsection{Directionality}
\label{sec:Directionality}

Translation quality is also affected by the nature of the source text. 
In this respect, we note that from the 2,001 sentences in the WMT 2017 Chinese--English test set, half were originally written in Chinese; the remaining half were originally written in English and then manually translated into Chinese. This Chinese reference file (half original, half translated) was then manually translated into English by \citet{hassan2018achieving} to make up the reference for assessing human--machine parity. Therefore, 50\percent of the reference comprises direct English translations from the original Chinese, while 50\percent are English translations from the human-translated file from English into Chinese, \ie*, backtranslations of the original English.

According to \citet{Laviosa1998}, translated texts differ from their originals in that they are simpler, more explicit, and more normalised. For example, the synonyms used in an original text may be replaced by a single translation. These differences are referred to as translationese, and have been shown to affect translation quality in the field of machine translation \citep{kurokawa2009automatic,daems2017translationese, toral-2019-post, castilho-etal-2019-influences}. 

We test whether translationese has an effect on assessing parity between translations produced by humans and machines, using relative rankings of translations in the WMT 2017 Chinese--English test set by five raters (see \Section{Raters}). Our hypothesis is that the difference between human and machine translation quality is smaller when source texts are translated English (translationese) rather than original Chinese, because a translationese source text should be simpler and thus easier to translate for an MT system. We confirm Laviosa's observation that “translationese” Chinese (that started as English) exhibits less lexical variety than “natively” Chinese text and demonstrate that translationese source texts are generally easier for MT systems to score well on.

\begin{table}
    \centering
    \begin{tabular}{llrlrlr}
\toprule
\textbf{Rank} & \multicolumn{6}{l}{\textbf{Original Language}} \\
              & \multicolumn{2}{l}{Both} & \multicolumn{2}{l}{Chinese} & \multicolumn{2}{l}{English} \\[-0.5em]
              & \multicolumn{2}{l}{$n=6675$} & \multicolumn{2}{l}{$n=3873$} & \multicolumn{2}{l}{$n=2802$} \\ 
              \hline
1		 & \HA & 1.587~*	    & \HA & 1.939~*        & \MT & 1.059~\phantom{*} \\
2		 & \MT & 1.231~*	        & \MT & 1.199~*	            & \HA & 0.772~* \\
3		 & \GG & -2.819~\phantom{*}	& \GG & -3.144~\phantom{*}  & \GG & -1.832~\phantom{*} \\
\bottomrule
\end{tabular}
    \smallskip
    \caption{Ranks of the translations given the original language of the source side of the test set shown with their TrueSkill score (the higher the better). An asterisk next to a translation indicates that this translation is significantly better than the one in the next rank at $p\leq.05$.}
    \label{tab:orig_lang}
\end{table}

Table \ref{tab:orig_lang} shows the TrueSkill scores for translations (\HA*, \MT*, and \GG*) of the entire test set (Both) versus only the sentences originally written in Chinese or English therein. The human translation \HA outperforms the machine translation \MT significantly when the original language is Chinese, while the difference between the two is not significant when the original language is English (\ie*, translationese input).

We also compare the two subsets of the test set, original and translationese, using type-token ratio (TTR). 
Our hypothesis is that the TTR will be smaller for the translationese subset, thus its simpler nature getting reflected in a less varied use of language.
While both subsets contain a similar number of sentences (1,001 and 1,000), the Chinese subset contains more tokens (26,468) than its English counterpart (22,279). We thus take a subset of the Chinese (840 sentences) containing a similar amount of words to the English data (22,271 words). We then calculate the TTR for these two subsets using bootstrap resampling. The TTR for Chinese ($M=0.1927$, $SD=0.0026$, 95\percent confidence interval $[0.1925,0.1928]$) is 13\percent higher than that for English ($M=0.1710$, $SD=0.0025$, 95\percent confidence interval $[0.1708,0.1711]$).

Our results show that using translationese (Chinese translated from English) rather than original source texts results in higher scores for MT systems in human evaluation, and that the lexical variety of translationese is smaller than that of original text.

\section{Recommendations}
\label{sec:Recommendations}

Our experiments in Sections~\ref{sec:Raters}--\ref{sec:ReferenceTranslations} show that machine translation quality has not yet reached the level of professional human translation, and that human evaluation methods which are currently considered best practice fail to reveal errors in the output of strong NMT systems. In this section, we recommend a set of evaluation design changes that we believe are needed for assessing human--machine parity, and will strengthen the human evaluation of MT in general.

\paragraph{(R1) Choose professional translators as raters.} In our blind experiment (\Section{Raters}), non-experts assess parity between human and machine translation where professional translators do not, indicating that the former neglect more subtle differences between different translation outputs. 

\paragraph{(R2) Evaluate documents, not sentences.} When evaluating sentences in random order, professional translators judge machine translation more favourably as they cannot identify errors related to textual coherence and cohesion, such as different translations of the same product name. Our experiments show that using whole documents (\ie*, full news articles) as unit of evaluation increases the rating gap between human and machine translation (\Section{Context}).

\paragraph{(R3) Evaluate fluency in addition to adequacy.} Raters who judge target language fluency without access to the source texts show a stronger preference for human translation than raters with access to the source texts (Sections~\ref{sec:Context} and~\ref{sec:Quality}). In all of our experiments, raters prefer human translation in terms of fluency while, just as in \citeg{hassan2018achieving} evaluation, they find no significant difference between human and machine translation in sentence-level adequacy (Tables~\ref{tab:mt-vs-a} and~\ref{tab:mt-vs-b}).
Our error analysis in \Table{quality_qualitative} also indicates that MT still lags behind human translation in fluency, specifically in grammaticality.

\paragraph{(R4) Do not heavily edit reference translations for fluency.} In professional translation workflows, texts are typically revised with a focus on target language fluency after an initial translation step. As shown in our experiment in \Section{Quality}, aggressive revision can make translations more fluent but less accurate, to the degree that they become indistinguishable from MT in terms of accuracy (\Table{mt-vs-b}).

\paragraph{(R5) Use original source texts.} Raters show a significant preference for human over machine translations of texts that were originally written in the source language, but not for source texts that are translations themselves (\Section{Directionality}). Our results are further evidence that translated texts tend to be simpler than original texts, and in turn easier to translate with MT.

Our work empirically strengthens and extends the recommendations on human MT evaluation in previous work \citep{Laeubli2018,Toral2018}, some of which have meanwhile been adopted by the large-scale evaluation campaign at WMT 2019 \citep{WMT2019}: the new evaluation protocol uses original source texts only (R5) and gives raters access to document-level context (R2). The findings of WMT 2019 provide further evidence in support of our recommendations. In particular, human English to Czech translation was found to be significantly better than MT \citep[p.~28]{WMT2019}; the comparison includes the same MT system (\texttt{CUNI-Transformer-T2T-2018}) which outperformed human translation according to the previous protocol \citep[p.~291]{WMT2018}. Results also show a larger difference between human translation and MT in document-level evaluation.\footnote{Specifically, the absolute difference between HUMAN and \texttt{CUNI-Transformer-T2T-2018} in terms of average
standardized human scores is 11--22\% for segment-level evaluation, 24\% for segment-level evaluation with document-level context, and 39\% for document-level evaluation \citep[p.~28]{WMT2019}.}

We note that in contrast to WMT, the judgements in our experiments are provided by a small number of human raters: five in the experiments of Sections~\ref{sec:Raters} and~\ref{sec:Directionality}, four per condition (adequacy and fluency) in \Section{Context}, and one in the fine-grained error analysis presented in \Section{Quality}. Moreover, the results presented in this article are based on one text domain (news) and one language direction (Chinese to English), and while a large-scale evaluation with another language pair supports our findings (see above), further experiments with more languages, domains, and raters will be required to increase their external validity.

\section{Conclusion}
\label{sec:Conclusion}

We compared professional human Chinese to English translations to the output of a strong MT system. 
In a human evaluation following best practices, \citet{hassan2018achieving} found no significant difference between the two, concluding that their NMT system had reached parity with professional human translation. 
Our blind qualitative analysis, however, showed that the machine translation output contained significantly more incorrect words, omissions, mistranslated names, and word order errors. 

Our experiments show that recent findings of human--machine parity in language translation are owed to weaknesses in the design of human evaluation campaigns. 
We empirically tested alternatives to what is currently considered best practice in the field, and found that the choice of raters, the availability of linguistic context, and the creation of reference translations have a strong impact on perceived translation quality. 
As for the choice of raters, professional translators showed a significant preference for human translation, while non-expert raters did not.
In terms of linguistic context, raters found human translation significantly more accurate than machine translation when evaluating full documents, but not when evaluating single sentences out of context. 
They also found human translation significantly more fluent than machine translation, both when evaluating full documents and single sentences.
Moreover, we showed that aggressive editing of human reference translations for target language fluency can decrease adequacy to the point that they become indistinguishable from machine translation, and that raters found human translations significantly better than machine translations of original source texts, but not of source texts that were translations themselves.

Our results strongly suggest that in order to reveal errors in the output of strong MT systems, the design of MT quality assessments with human raters should be revisited. 
To that end, we have offered a set of recommendations, supported by empirical data, which we believe are needed for assessing human--machine parity, and will strengthen the human evaluation of MT in general.
Our recommendations have the aim of increasing the validity of MT evaluation, but we are aware of the high cost of having MT evaluation done by professional translators, and on the level of full documents.
We welcome future research into alternative evaluation protocols that can demonstrate their validity at a lower cost.

\bibliographystyle{apacite}
\bibliography{references}
\end{CJK*}
\end{document}